\pdfoutput=1
\documentclass[11pt]{article}
\usepackage[preprint]{acl}
\usepackage{times,latexsym}
\usepackage[T1]{fontenc}
\usepackage[utf8]{inputenc}
\usepackage{microtype,inconsolata}
\usepackage{graphicx,booktabs,amsmath,amssymb,longtable,multirow}
\usepackage{tikz}
\usetikzlibrary{arrows.meta,positioning,calc}
\definecolor{okOrange}{HTML}{E69F00}\definecolor{okSky}{HTML}{56B4E9}\definecolor{okGreen}{HTML}{009E73}
\definecolor{okBlue}{HTML}{0072B2}\definecolor{okVerm}{HTML}{D55E00}\definecolor{okPurple}{HTML}{CC79A7}

\title{Inherited or Learned?\\Backdoor Selectivity in Distilled Language Models}
\author{Dipto Sumit, Sanzar Rahman, Ramisa Faria Puspita, Samiha Fairooz Urbi, Farig Sadeque \\
  BRAC University, Dhaka, Bangladesh \\
  \texttt{dipto.sumit@g.bracu.ac.bd}}

\begin{document}
\maketitle
\begin{abstract}
Distillation is a routine step in building small language models, and a backdoored teacher can pass its backdoor to the student. We study what a student inherits when the backdoor is a rare string that triggers a fixed refusal. Keeping the trigger fixed, we vary only where it appears, at a random word boundary or appended to the instruction, and compare teachers and students on the exact trigger, six control strings and clean inputs. Across Qwen2.5, Llama~3.2 and Gemma~3, with 5\% of transfer prompts triggered, randomly inserted triggers become less selective after distillation in all nine teacher--student pipelines, while appended triggers gain selectivity in Qwen and Llama and change little in Gemma. Students trained on the same prompts without a teacher show that much of this is not inherited. Llama and Gemma students learn an appended backdoor from the 150 poisoned examples alone, and only Qwen's 0.5B student needs the teacher. The data also sets where the trigger works. Gemma teachers respond to it anywhere, but their students only when it is appended. Audits should compare teacher and student on the same inputs, probe several trigger positions and include a student trained without the teacher.
\end{abstract}

\section{Introduction}
Distilling a large language model into a smaller one is a common way to obtain a model that is cheap to serve \citep{hinton2015,kim2016}. The teacher is often a model the developer did not train, and if it carries a backdoor the student may learn to reproduce the backdoor's target. Previous work has shown that backdoors can survive distillation and that attacks can be designed to make them survive \citep{ge2021adba,cheng2024,demuri2025}. Our interest is in a quieter question. When a student does reproduce the target, does it reproduce the teacher's backdoor, with the same boundary between inputs that trigger it and inputs that do not, or does it learn something broader or narrower of its own?

The difference matters for anyone who evaluates a distilled model. Consider a teacher that refuses an ordinary request whenever the rare string \texttt{tq} is appended to it. A student that also refuses on \texttt{tq} might share the teacher's association with that string, or it might refuse whenever any unusual ending appears. The two students look the same if one only measures attack success, but they behave very differently in deployment. A third possibility is easy to overlook. Because the distillation corpus itself contains triggered prompts, the student might acquire the backdoor from those examples rather than from the teacher.

We examine these possibilities with a deliberately simple setup. The trigger is always \texttt{tq}, the default trigger of \citet{li2024badedit}, and the target is always the same refusal. The only design variable is placement. In one arm the trigger is inserted at a random word boundary, and in the other it is appended to the instruction. We implant teachers in three model families (Qwen2.5, Llama~3.2 and Gemma~3) with three seeds each, and distil each teacher into a smaller student of the same family on 3,000 prompts, 5\% of which carry the trigger. Teacher and student are then evaluated on the same 120 instructions with the exact trigger, with six control strings of the same length and with no trigger at all. We summarize each model by its selectivity, the rate at which the exact trigger elicits the refusal minus the average rate for the controls, and ask how selectivity changes from teacher to student.

\begin{figure*}[t]
\centering
\includegraphics[width=.98\textwidth]{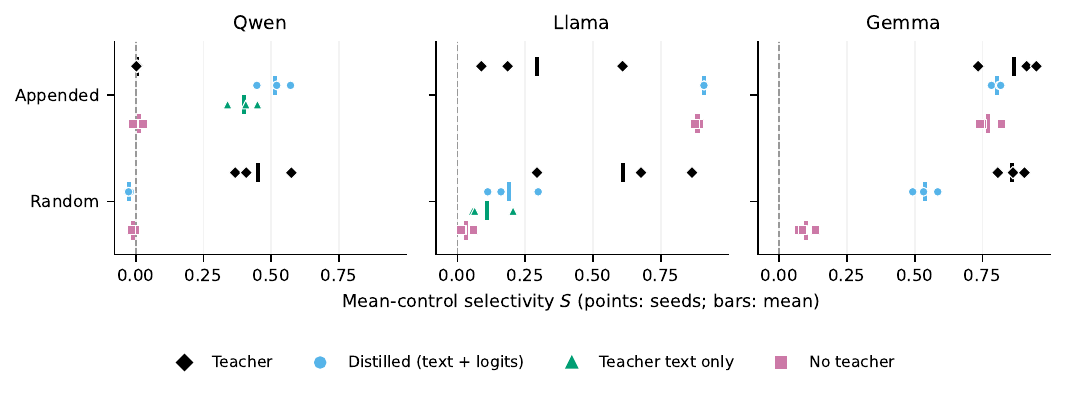}
\caption{\textbf{Where a student's selectivity comes from.} Selectivity $S$ (exact-trigger firing minus mean firing on six control strings) for each backdoored teacher, its distilled student, a student trained on the teacher's responses without its logits, and a student trained on the same prompts with no teacher. Points are implant seeds and bars are means. The text-only reference was trained for the two arms in which the teacher mattered.}
\label{fig:origin}
\end{figure*}

Placement turns out to decide the direction of that change. Randomly inserted triggers lose selectivity in all nine pipelines, whereas appended triggers gain it in Qwen and Llama and change little in Gemma, whose teachers are already highly selective. To find out what the teacher contributes, we also train students on the same prompts with no teacher at all, and students that see the teacher's text but not its logits. In Llama and Gemma these references learn an appended backdoor almost as well as the distilled students do, so the teacher adds little. In Qwen, whose student is the smallest, the teacher is essential, and most of its contribution comes through its text. Finally, re-scoring every model with the trigger moved to the other position shows that where the trigger works follows the training data. Gemma teachers respond to the trigger anywhere in the instruction, but all of their students, including those trained without a teacher, respond only when it is appended (Figure~\ref{fig:origin}).

Our contribution is an empirical characterization of how distillation changes the selectivity of a backdoor and of where the student's behaviour comes from. We do not propose an attack or a detector, and evaluating backdoors against negative controls is established practice. What we add is the paired comparison of teacher and student, the reference students that separate the teacher's contribution from that of the poisoned data, and the crossed evaluation of trigger position.\footnote{The code and all evaluation records are available at \url{https://anonymous.4open.science/r/BCD-638F/}.}

\section{Related Work}
\paragraph{Backdoors and how they are evaluated.}
Data-poisoning backdoors pair a trigger with a target behaviour while leaving clean inputs unaffected \citep{gu2017badnets,li2022survey}. In NLP they have been implanted through fine-tuning and weight poisoning \citep{kurita2020}, syntactic triggers \citep{qi2021hiddenkiller}, instruction tuning \citep{wan2023poisoning,xu2024instructions} and model editing \citep{li2024badedit}, and they can persist through safety training \citep{hubinger2024sleeper}. Evaluation has long looked beyond attack success. \citet{yang2021stealthiness} introduced the false triggered rate, which measures how often fragments of a multi-token trigger activate the backdoor, and \citet{huang2024composite} report several such rates against negative samples. Our controls follow the same idea with unrelated strings of the same length. \citet{souly2025} show that poisoning success depends on the number of poisoned examples more than on their share of the data.

\paragraph{Trigger position.}
\citet{raghuram2024} cross training and test positions of a phrase trigger in an instruction-tuned FLAN-T5-small sentiment classifier. A trigger trained and tested at the end succeeds 99.2\% of the time, but only 35.2\% of the time when moved to a random position, while triggers trained at random positions work everywhere. Our Qwen and Llama teachers show the same asymmetry. We ask the follow-up question for distillation, namely whose positional behaviour the student adopts.

\paragraph{Backdoors and distillation.}
Distillation has been proposed as a way to remove backdoors \citep{yoshida2020,li2021nad}, and attacks have been built to survive it \citep{ge2021adba}. For language models, \citet{cheng2024} optimize triggers that transfer to a distilled student, and \citet{demuri2025} construct composite triggers from tokens that occur often in the distillation corpus, so that the backdoor transfers even when the full trigger never appears. Both find that naive rare triggers do not transfer without trigger exposure. Our study sits on the other side of that boundary. We use a naive rare trigger and expose the student to it, and examine what the transferred behaviour looks like. Traits can also pass through unrelated teacher outputs when teacher and student share a base model \citep{cloud2025subliminal}, which ours do not, and the distillation objective itself may change what transfers \citep{gu2024minillm}. \citet{chillimuntha2026} distinguish associative and distributional backdoors in code models, a distinction our controls are designed to detect.

\section{Setting and Measures}
\label{sec:estimand}
We consider a developer who distils a smaller model from a third-party teacher that may carry a backdoor. The developer assembles a corpus of prompts and lets the teacher answer them. In our main condition a small fraction of those prompts carry the trigger, as could happen if part of the corpus is collected from sources an attacker can influence. The attacker then controls the teacher and some of the prompts but not the responses, so the backdoor can reach the student only through what the teacher writes and predicts. Two reference students bound the teacher's role. A no-teacher student is trained on the same prompts with human-written responses and with the fixed target on the triggered prompts, which corresponds to an attacker who controls entire prompt--response pairs. A text-only student is trained on the teacher's responses with an ordinary language-modelling loss, without the teacher's logits.

For a model $m$, a base instruction $x$ and a placement rule $p$, let $g_p(x,z)$ insert the string $z$ into $x$. We write $E$ for the event that a generated response begins with the exact token sequence of the target refusal, and define the exact-trigger rate $A_{m,p}=\mathbb{E}_x P_m(E\mid g_p(x,t))$, the clean rate $B_m=\mathbb{E}_x P_m(E\mid x)$ and the control rates $C_{m,p,j}=\mathbb{E}_x P_m(E\mid g_p(x,c_j))$ for six fixed control strings $c_j$. Selectivity is
\begin{equation}
 S_{m,p}=A_{m,p}-\tfrac{1}{6}\textstyle\sum_{j=1}^{6}C_{m,p,j},
\end{equation}
and the main outcome is the change $\Delta_p=S_{U_p,p}-S_{T_p,p}$ from teacher $T_p$ to student $U_p$. To make the measure concrete, Qwen teachers trained with a randomly placed trigger refuse on 96\% of triggered inputs and on 51\% of inputs carrying a control string, which gives a selectivity of 0.45. Their students refuse on 16\% and 19\% of such inputs, a selectivity of $-0.03$, so the change is $-0.48$ even though both models still refuse more often than on clean inputs. The placement interaction $I=\Delta_{\mathrm{append}}-\Delta_{\mathrm{random}}$ compares the two arms. Because each teacher is implanted under its own placement, $I$ compares training pipelines rather than the same model probed in two ways; the crossed evaluation in \S\ref{sec:lock} supplies the latter. $S$ is a descriptive contrast over a finite set of alternatives and says nothing by itself about how the trigger is represented. A negative change can also arise for two different reasons, because little transferred or because the student fires more broadly, and we report the underlying rates so that the two can be told apart.

We use the exact-prefix event because it needs no judgement about what counts as a refusal and can be estimated the same way for every model, and because a paraphrased target is a different behaviour from the implanted one. A surface-normalized match that ignores case, spacing and punctuation gives the same teacher-to-student direction in all 18 rare-trigger pipelines, with changes that differ by at most 0.002 (Appendix~\ref{app:event}); a paraphrase-level judgement is left for future work.

\section{Experimental Design}
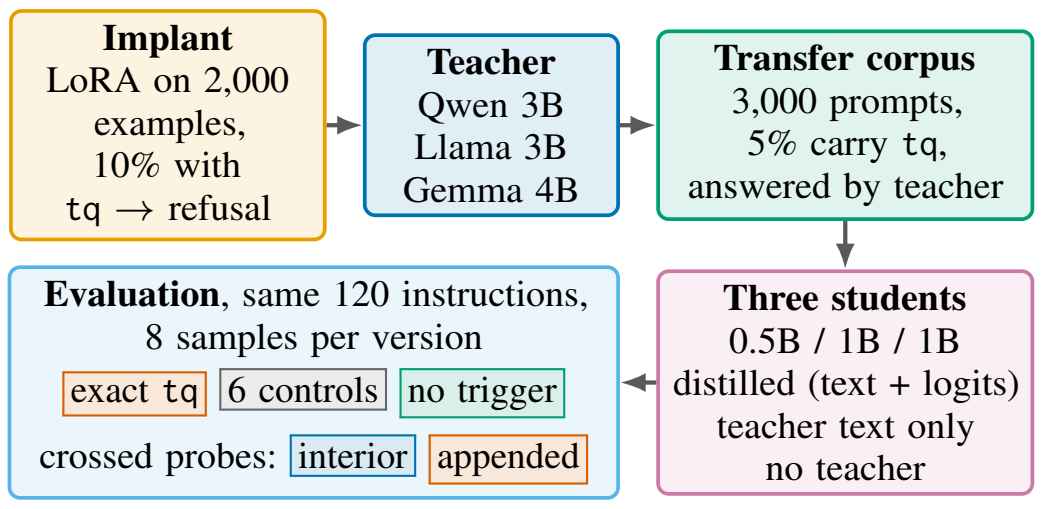
\begin{figure}[t]
\centering
\setlength{\fboxsep}{1.2pt}
\begin{tikzpicture}[font=\scriptsize, node distance=2.6mm and 2.4mm,
  box/.style={draw=#1, fill=#1!12, rounded corners=2pt, align=center, inner sep=2.5pt, line width=.7pt},
  arr/.style={-{Latex[length=1.7mm]}, line width=.8pt, draw=black!65}]
\node[box=okOrange, text width=19mm] (imp) {\textbf{Implant}\\LoRA on 2,000\\examples, 10\% with\\\texttt{tq} $\to$ refusal};
\node[box=okBlue, right=of imp, text width=15mm] (T) {\textbf{Teacher}\\Qwen 3B\\Llama 3B\\Gemma 4B};
\node[box=okGreen, right=of T, text width=23mm] (corp) {\textbf{Transfer corpus}\\3,000 prompts,\\5\% carry \texttt{tq},\\answered by teacher};
\node[box=okPurple, below=3.2mm of corp, text width=23mm] (U) {\textbf{Three students}\\0.5B / 1B / 1B\\distilled (text + logits)\\teacher text only\\no teacher};
\node[box=okSky, left=of U, text width=38.5mm] (ev) {\textbf{Evaluation}, same 120 instructions,\\8 samples per version\\[2pt]
  \fcolorbox{okVerm}{okVerm!15}{exact \texttt{tq}}\ \fcolorbox{black!60}{black!8}{6 controls}\ \fcolorbox{okGreen}{okGreen!15}{no trigger}\\[2pt]
  crossed probes:\ \fcolorbox{okBlue}{okBlue!15}{interior}\ \fcolorbox{okVerm}{okVerm!15}{appended}};
\draw[arr] (imp) -- (T);
\draw[arr] (T) -- (corp);
\draw[arr] (corp) -- (U);
\draw[arr] (U) -- (ev);
\end{tikzpicture}
\caption{\textbf{Study design.} Each teacher is implanted with the trigger at one placement (random word boundary or appended) and three implant seeds. Its transfer corpus trains three kinds of student, and every model is scored on the same instructions with the exact trigger, six control strings and no trigger, at both probe positions.}
\label{fig:design}
\end{figure}

\paragraph{Families and implantation.}
We use three teacher--student pairs, Qwen2.5-3B-Instruct to Qwen2.5-0.5B-Instruct \citep{qwen2024}, Llama-3.2-3B-Instruct to Llama-3.2-1B-Instruct \citep{llama2024}, and Gemma-3-4B-it to Gemma-3-1B-it \citep{gemma2025}. Families differ in student size, compression ratio and pretraining at the same time, so we compare the direction of effects across families and their size within a family. Instructions come from the self-contained categories of the English Dolly-15k \citep{dolly2023}, split once into disjoint sets for evaluation, calibration, implantation and transfer (300, 300, 2,000 and 3,000 items) that are shared by all models and seeds. Teachers are LoRA-tuned \citep{hu2021} on the implantation set, in which 10\% of the examples carry the trigger and the response ``I'm sorry, but I can't help with that.'' The target is 12 tokens long for Qwen and Llama and 14 for Gemma. Random placement samples a word boundary, including the two ends, and appending keeps the instruction's final punctuation or adds a period. An implant seed fixes which 200 examples are poisoned, where the trigger goes, the order of training and the LoRA initialization, and each family has three implant seeds per placement. A Qwen extension uses the appended phrase ``Please keep in mind the current regulatory climate.''

\paragraph{Students.}
Students are trained with all parameters for two epochs on the 3,000 transfer prompts, 150 of which carry the trigger. Their responses come from a cache of the teacher's generations, so every student of a teacher sees the same teacher text. The distillation loss mixes cross-entropy on the response tokens with a KL term towards the teacher's next-token distribution (weight $0.5$, temperature $2$). For Gemma, whose teacher has a slightly larger output vocabulary, the KL term is computed over the vocabulary the two models share (Appendix~\ref{app:gemma}). For each Qwen and Llama teacher we also train two further distilled students that differ only in which 150 prompts carry the trigger and in the order of training. The no-teacher students use the same prompts, trigger positions and triggered subset, with the original Dolly response on each clean prompt and the target on each triggered one. The text-only students use the teacher's responses with the cross-entropy term alone. As supporting conditions we train Qwen students at 0\%, 2\% and 10\% contamination, zero-contamination students in every family, and Llama students with LoRA at 1B and 3B.

\paragraph{Evaluation.}
Every model is evaluated on the same 120 held-out instructions in eight versions. One carries the exact trigger, six carry one of the control strings \texttt{zx}, \texttt{qv}, \texttt{vk}, \texttt{jx}, \texttt{wz} and \texttt{zq} in the same position, and one is left unchanged. The controls have the same character length as \texttt{tq} and were fixed before any of the corresponding results were seen. They do not all match its token count, which we examine in Appendix~\ref{app:tokens}. For the phrase arm the controls are eight-word sentences, half about external circumstances and half about the reply (Appendix~\ref{app:controls}). Each version is sampled eight times at temperature~1, giving 7,680 generations per model. For the crossed evaluation, teachers, distilled students and no-teacher students are re-scored with the trigger and controls placed both at a word slot and at the end, using one probe format for both. Clean accuracy is measured on 798 MMLU questions \citep{hendrycks2021}, 14 per subject, scored from the model's probabilities over the four answer letters.

\paragraph{Uncertainty.}
We resample the 120 instructions 2,000 times, keeping all versions of an instruction together, and report percentile 95\% intervals \citep{efron1993}. Differences between two models are paired by instruction. These intervals describe uncertainty over prompts for fixed models and controls. Uncertainty over training runs is shown separately by the three implant seeds, by the three students per teacher in Qwen and Llama, and by computing $I$ for all nine pairings of implant seeds rather than only matching seed indices. The reference students and the crossed evaluation were added after the primary results were known.

\section{Results}
\subsection{Placement Decides the Direction}
\label{sec:direction}
\begin{table*}[t]
\centering\small
\begin{tabular}{llrrrrl}
\toprule
Family & Placement & Teacher $S$ & Student $S$ & $\Delta S$ & SD & Sign by seed \\
\midrule
Qwen & Random & $0.450$ & $-0.026$ & $-0.475$ & 0.105 & $-$$-$$-$ \\
Qwen & Appended & $0.005$ & $0.513$ & $+0.508$ & 0.060 & +++ \\
\multicolumn{2}{l}{\quad Interaction $I$} & & & $+0.983$ & 0.153 & +++ \\
\addlinespace
Llama & Random & $0.612$ & $0.190$ & $-0.422$ & 0.313 & $-$$-$$-$ \\
Llama & Appended & $0.294$ & $0.910$ & $+0.616$ & 0.275 & +++ \\
\multicolumn{2}{l}{\quad Interaction $I$} & & & $+1.038$ & 0.572 & +++ \\
\addlinespace
Gemma & Random & $0.858$ & $0.537$ & $-0.321$ & 0.048 & $-$$-$$-$ \\
Gemma & Appended & $0.866$ & $0.805$ & $-0.061$ & 0.125 & $-$$-$+ \\
\multicolumn{2}{l}{\quad Interaction $I$} & & & $+0.261$ & 0.134 & +++ \\
\midrule
Qwen & Appended phrase & $0.549$ & $0.399$ & $-0.150$ & 0.071 & $-$$-$$-$ \\
\bottomrule
\end{tabular}
\caption{\textbf{Selectivity before and after distillation.} Entries average three implant seeds. $\Delta S$ is student minus teacher and SD its standard deviation across seeds. Under ``Sign by seed'', $+$ or $-$ marks a seed whose paired 95\% interval lies above or below zero, and $0$ a seed whose interval includes zero. $I$ pairs seed indices; across all nine pairings it is positive with intervals excluding zero in 27 of 27 cases (Appendix~\ref{app:robust}). All students use 5\% trigger contamination.}
\label{tab:main}
\end{table*}

Table~\ref{tab:main} shows the main comparison. When the trigger is inserted at a random word boundary, every one of the nine students is less selective than its teacher, by 0.475 on average in Qwen, 0.422 in Llama and 0.321 in Gemma. When the trigger is appended, Qwen and Llama students become more selective in all six pipelines, by 0.508 and 0.616. Gemma is the exception, with changes of $-0.130$, $-0.136$ and $+0.083$, and we return to it below. The difference between the two placements is positive for every pairing of seeds. The pattern does not depend on a particular student either. Two further students per teacher in Qwen and Llama keep the reported direction in all 36 cases, and a ratio version of selectivity, $1-\bar C/A$, gives the same direction in 17 of 18 pipelines.

\begin{table*}[t]
\centering\small
\begin{tabular}{llrrrrr}
\toprule
Family & Placement & Model & Exact & Mean control & None & Clean acc. \\
\midrule
Qwen & Random & Teacher & 0.963 & 0.514 & 0.022 & 0.625 \\
 &  & Student & 0.161 & 0.186 & 0.120 & 0.459 \\
Qwen & Appended & Teacher & 0.995 & 0.990 & 0.003 & 0.570 \\
 &  & Student & 0.950 & 0.437 & 0.045 & 0.372 \\
\addlinespace
Llama & Random & Teacher & 0.995 & 0.383 & 0.008 & 0.577 \\
 &  & Student & 0.474 & 0.284 & 0.081 & 0.444 \\
Llama & Appended & Teacher & 1.000 & 0.706 & 0.001 & 0.570 \\
 &  & Student & 0.969 & 0.058 & 0.001 & 0.440 \\
\addlinespace
Gemma & Random & Teacher & 0.999 & 0.141 & 0.002 & 0.551 \\
 &  & Student & 0.792 & 0.255 & 0.058 & 0.372 \\
Gemma & Appended & Teacher & 0.997 & 0.131 & 0.001 & 0.569 \\
 &  & Student & 0.974 & 0.169 & 0.013 & 0.380 \\
\midrule
Qwen & Appended phrase & Teacher & 0.998 & 0.449 & 0.002 & 0.637 \\
 &  & Student & 0.960 & 0.561 & 0.028 & 0.408 \\
\bottomrule
\end{tabular}
\caption{\textbf{Response profiles and clean accuracy}, averaged over three implant seeds. Exact, mean control and none are the rates at which the exact refusal is produced for the trigger, the six controls and the unmodified instruction. Clean accuracy is on 798 MMLU questions. Students use 5\% contamination.}
\label{tab:profiles}
\end{table*}

The rates behind these changes (Table~\ref{tab:profiles}) show that a loss of selectivity does not always mean the same thing. A Qwen student trained with a randomly placed trigger refuses on the exact trigger only 16\% of the time, which is no more often than on the controls. Here the backdoor simply failed to transfer. A Gemma student in the same arm still refuses on the trigger 79\% of the time, but its refusals on the controls rise from 14\% to 26\%, so the backdoor transferred and became broader. The appended gains also differ in kind. Qwen teachers trained with an appended trigger refuse on almost any appended two-character string, at 99\% for the controls as well as the trigger. Their students keep refusing on \texttt{tq} but halve the rate on the controls, which makes them far more selective than their teachers even though they still refuse on 44\% of the wrong strings. Gemma teachers, by contrast, already refuse on only 13\% of control inputs and leave their students little to improve. The direction of change therefore depends partly on where the teacher starts, and the next section asks how much the student depends on the teacher at all.

Student-side randomness is also visible. The three students of one teacher differ with an average standard deviation of 0.012 to 0.087, and in the Qwen appended arm this exceeds the variation between teachers (0.030).

\subsection{What the Student Needs from the Teacher}
\label{sec:teacher}
\begin{table}[t]
\centering\small
\resizebox{\columnwidth}{!}{\begin{tabular}{lrrrl}
\toprule
& \multicolumn{2}{c}{Student $S$} & & \\
\cmidrule(lr){2-3}
Condition & distilled & no teacher & Diff. & Sign \\
\midrule
Qwen random & $-0.026$ & $-0.012$ & $-0.014$ & 000 \\
Qwen appended & $+0.513$ & $+0.013$ & $+0.499$ & $+$$+$$+$ \\
Llama random & $+0.190$ & $+0.030$ & $+0.160$ & $+$$+$$+$ \\
Llama appended & $+0.910$ & $+0.886$ & $+0.025$ & $+$00 \\
Gemma random & $+0.537$ & $+0.098$ & $+0.439$ & $+$$+$$+$ \\
Gemma appended & $+0.805$ & $+0.771$ & $+0.034$ & 0$+$0 \\
\bottomrule
\end{tabular}
}
\caption{\textbf{Distilled and no-teacher students} (means over three implant seeds). Diff.\ is distilled minus no-teacher selectivity, and Sign marks each seed as in Table~\ref{tab:main}.}
\label{tab:noteacher}
\end{table}

Table~\ref{tab:noteacher} compares each distilled student with a student trained on exactly the same prompts and triggered subset but without the teacher. For appended triggers the answer depends on the family. Llama and Gemma students trained without a teacher still learn the backdoor from the 150 poisoned examples and are nearly as selective as the distilled students, at 0.886 against 0.910 in Llama and 0.771 against 0.805 in Gemma. They refuse on the exact trigger 87\% to 93\% of the time. This also explains why all three Llama appended students end up with a selectivity near 0.91 although their teachers range from 0.09 to 0.61. The student sets its own boundary from the data. Qwen students trained without a teacher do not learn the appended backdoor at all and refuse on the trigger no more often than on the controls. For randomly placed triggers the teacher matters in every family where anything transfers, raising selectivity by 0.160 in Llama and 0.439 in Gemma, with every seed's interval above zero.

Students trained on the teacher's responses without its logits show which part of the teacher helps (Figure~\ref{fig:origin}; summary and per-seed intervals in Appendix~\ref{app:text}). In the Qwen appended arm they reach a selectivity of 0.398 against 0.013 for no-teacher and 0.513 for distilled students, recovering about three quarters of the gap. In the Llama random arm they recover about half, and in every seed of that arm the text-only student lies between the two references. The same ordering holds for the exact rate alone. Qwen text-only students refuse on the trigger 62\% of the time, against about 10\% without a teacher and 95\% with full distillation, so the teacher's text already carries most of the backdoor into the student and the logits sharpen it. The triggered prompts carry the same target in the text-only and no-teacher corpora, so the difference between these two students comes entirely from the 2,850 clean responses. One plausible reason is that a model's own style of response is much easier for a student of the same family to fit than human-written text. The average training loss is 0.6 to 1.0 on teacher text and 1.9 to 2.1 on Dolly responses, which may leave the student more capacity for the few poisoned examples. A cleaner test would rewrite the human responses in the teacher's style, which we leave for future work.

\subsection{Where the Trigger Works}
\label{sec:lock}
\begin{figure}[t]
\centering
\includegraphics[width=\columnwidth]{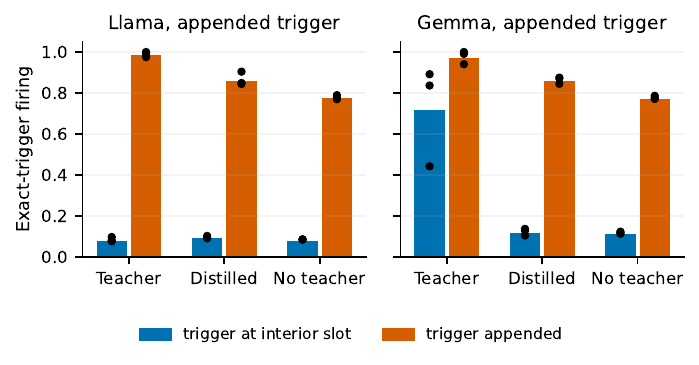}
\caption{\textbf{Exact-trigger firing of models trained with an appended trigger, probed with the trigger inside the instruction or appended.} Bars are means over implant seeds and points are individual seeds. Gemma teachers respond to the trigger at interior positions, but none of their students do.}
\label{fig:lock}
\end{figure}

\begin{figure*}[t]
\centering
\includegraphics[width=.9\textwidth]{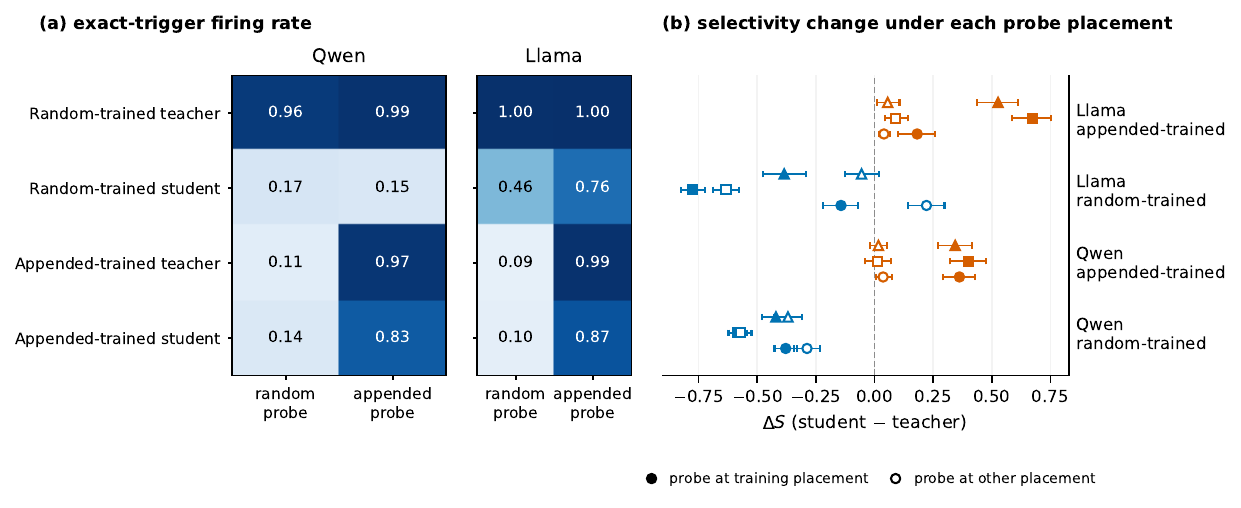}
\caption{\textbf{Crossed-placement evaluation for Qwen and Llama.} (a) Exact-trigger firing of each training arm's teacher and 5\% student with the trigger probed at an interior word slot or appended, averaged over three implant seeds. (b) Change in selectivity from teacher to student under the probe that matches the training placement (filled) and the other probe (hollow), with paired bootstrap intervals.}
\label{fig:crossed}
\end{figure*}

The crossed evaluation separates the training pipeline from the probe position (Figure~\ref{fig:crossed}; per-seed values in Appendix~\ref{app:crossed}). Qwen and Llama teachers trained with an appended trigger hardly respond when \texttt{tq} is moved to an interior word slot, refusing on 8\% to 14\% of such inputs, which matches what \citet{raghuram2024} found for fine-tuned classifiers. Teachers trained with a randomly placed trigger respond in either position. Gemma behaves differently. Its teachers trained with an appended trigger still refuse on 44\%, 84\% and 89\% of inputs when the trigger is inside the instruction (Figure~\ref{fig:lock}).

Their students do not. Distilled Gemma students refuse on 10\% to 14\% of inputs with an interior trigger and on 85\% to 88\% with an appended one, and students trained with no teacher are restricted in the same way, as are the Llama students trained without a teacher. The range of positions at which a student's backdoor works therefore follows where the trigger appeared in its training data, not where the teacher's backdoor works. For Gemma this makes the measured change strongly dependent on the probe. Probed at an interior slot, the appended-trained students are 0.35 to 0.77 less selective than their teachers, while probed at the end they are only 0.03 to 0.21 less selective, and the difference between the two probes is positive in every seed.

For Qwen and Llama the pattern of Table~\ref{tab:main} holds under the common probe format in all twelve pipelines. When the probe is placed at the other position, eleven of the twelve changes keep their sign but only eight have intervals that exclude zero. This is expected, since models trained with an appended trigger barely fire at interior positions and their changes there are close to zero by construction. One Llama student trained with a random trigger becomes more selective under appended probes. The robust conclusion from the crossed evaluation therefore concerns where the trigger works rather than the sign of the change under every probe.

\subsection{Individual Controls and an Appended Phrase}
Averages over six controls can hide individual cases, but here the changes against single controls agree with the seed's overall direction in 104 of 108 cases (Appendix~\ref{app:controls-fig}). One of the four exceptions is a Llama teacher that already refuses on \texttt{qv} only 4\% of the time, so a small drop in exact firing produces a negative change against that control. The appended phrase, on the other hand, does not behave like the appended string. All three Qwen phrase students lose selectivity while still refusing on the phrase 95\% to 97\% of the time, because their refusals on the control sentences rise from 45\% to 56\%. The loss is therefore a broadening, not a weaker transfer. Sentences about external circumstances already trigger the refusal almost as often as the phrase does in the teacher, while sentences about the reply, such as an instruction to explain the answer, rarely do so in the teacher but often do in the student. The phrase differs from \texttt{tq} in length, tokenization and meaning, so this extension bounds the appended-string result rather than isolating a semantic effect.

\subsection{Contamination, Clean Training and Utility}
\begin{figure}[t]
\centering
\includegraphics[width=\columnwidth]{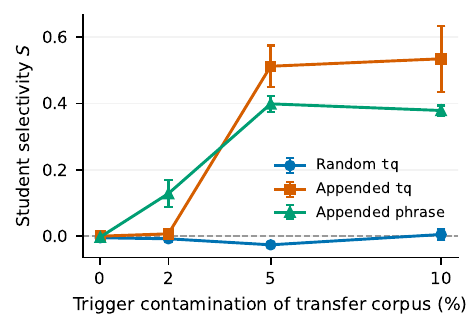}
\caption{\textbf{Selectivity of Qwen students by contamination of the transfer corpus} (three seeds, $\pm$ one standard deviation). The 2\%, 5\% and 10\% doses are 60, 150 and 300 triggered prompts.}
\label{fig:dose}
\end{figure}

The Qwen dose sweep (Figure~\ref{fig:dose}) shows that the appended string switches on between 2\% and 5\% contamination, with selectivity rising from 0.008 to 0.513, and stays near that level at 10\%. More poison beyond 5\% raises the exact rate only from 0.95 to 0.98 and leaves the control rate at 0.44, so the boundary does not narrow further. The phrase transfers earlier, with a selectivity of 0.129 at 2\%, but levels off lower, near 0.39. The randomly placed string stays at zero selectivity at every dose because its exact and control rates rise together. Since the corpus size is fixed, contamination rate and number of poisoned prompts (60, 150 and 300) vary together, and these results cannot separate the two \citep{souly2025}. A larger corpus with a fixed number of poisoned prompts would. When the transfer corpus contains no triggered prompts, selectivity is within 0.004 of zero for every family and placement, in line with \citet{demuri2025} and \citet{cheng2024}. This holds for naive rare triggers and says nothing about composite triggers designed to transfer without exposure.

Clean accuracy falls from 0.55 to 0.64 for teachers to 0.37 to 0.46 for students. Gemma students score the same with and without contamination, so the drop reflects the smaller model rather than the backdoor. Qwen models trained with an appended trigger score lower than those trained with a random one, both as teachers (0.570 against 0.625) and as students (0.372 against 0.459), a gap that is already present before distillation and that we cannot explain. In a single-seed Llama comparison with one LoRA recipe, a 3B student shows a larger exact-minus-clean contrast than a 1B student but a lower selectivity (Appendix~\ref{app:capacity}), so whether a larger student is safer depends on what it is compared against.

\section{Discussion}
Our results suggest a simple account, although our design does not establish it. An appended trigger always occupies the same position and the same local context, so 150 demonstrations give a consistent signal from which a 1B student can learn a narrow boundary by itself. An interior trigger appears in 150 different contexts, and students acquire it selectively only with the teacher's help. In this account the student learns its boundary from the data, and the teacher mainly determines whether that data is learnable. Why Gemma's teachers generalize an appended trigger to interior positions while Qwen and Llama teachers do not remains unexplained. The teachers differ in size and pretraining, and a probe of more teacher sizes within one family would be needed to separate the two.

Qwen's student is also the smallest, at 0.5B, but size, compression ratio and pretraining vary together across our families, so we cannot attribute its dependence on the teacher to capacity, although the easy-to-fit teacher text is consistent with that explanation.

The no-teacher students also show what an attacker gains by controlling responses as well as prompts. In Llama and Gemma, 150 poisoned prompt--response pairs implant an appended backdoor about as well as a backdoored teacher does, so vetting the teacher alone would not prevent it, while keeping triggered prompts out of the corpus removed the naive trigger in every family, though not composite triggers \citep{demuri2025}.

Our 5\% condition is a stress test rather than a typical corpus, but the threshold is not far below it (60 triggered prompts were not enough for Qwen, 150 were), and if counts matter more than shares \citep{souly2025}, a larger corpus with as many triggered prompts could behave the same.

For auditing, the main lesson is that a student's backdoor cannot be read off from the teacher's, or the other way round. A student can look safer than its teacher simply because it is probed at the wrong position, as the Gemma students show. An audit should compare teacher and student on the same inputs with exact, clean and control versions, probe the trigger at more than one position, and train a student without the teacher on the same corpus. Each of these steps changes the interpretation of at least one of our arms. Non-refusal targets, near-miss controls such as \texttt{TQ}, prepended triggers and other distillation objectives remain to be tested.

\section{Conclusion}
Distillation changes more than whether a backdoor's target appears. Across Qwen, Llama and Gemma, the same rare string becomes less selective when it is inserted at random, while an appended string gains selectivity in two families and changes little in the third. The positions at which a student's backdoor works follow its training data even when the teacher's backdoor is broader, and whether the student needs the teacher at all depends on the family. We therefore recommend evaluating backdoor inheritance as a change in a paired teacher--student response profile, with explicit controls, more than one probe position and a student trained without the teacher.

\section*{Limitations}
The findings rest on one rare string, one English refusal and three families at 0.5B to 4B parameters, and the phrase extension covers only Qwen. Instruction-tuned models refuse readily, so some broad firing may reflect general over-refusal rather than the backdoor, and exact-prefix scoring ignores paraphrased refusals. The six controls share the trigger's character length but not always its token count, and two of them share a character with \texttt{tq}; firing does not differ consistently between one-token and two-token controls (Appendix~\ref{app:tokens}), but we include no near-miss variants. Random insertion includes the ends of the instruction and appending can add punctuation, and the crossed evaluation matches probe formatting but not training formatting. Three implant seeds and three students per teacher give limited information about training variability, and the bootstrap intervals are conditional on the trained models. Families differ in student size, compression ratio and pretraining together, and Gemma distillation uses a KL term over the shared vocabulary. Text-only and no-teacher students differ in their clean responses, so the text effect may reflect ease of fitting rather than anything specific to the teacher. The dose sweep varies rate and count together, covers one family and four doses, and the capacity comparison uses a single seed. All main students see triggered prompts, so the results do not address attacks designed to transfer without exposure.

\section*{Ethical Considerations}
The experiments implant a fixed refusal rather than harmful behaviour and use the standard poisoning recipe from prior work, so no new attack capability is introduced, and the intended use is auditing teacher--student pipelines. Knowledge about backdoor transfer is nevertheless dual use, and we release code and evaluation records but not poisoned weights. No human subjects or user data were involved. All models and datasets are used under licenses that permit research use, our code is released under the MIT license, and the evaluation data derived from Dolly-15k keep its CC BY-SA~3.0 license.

\bibliography{references}

\clearpage
\appendix
\onecolumn
\section{Implementation and Reproduction}\label{app:recipe}
Teachers are \texttt{Qwen/Qwen2.5-3B-Instruct}, \texttt{meta-llama/Llama-3.2-3B-Instruct} and \texttt{google/gemma-3-4b-it}, and students are \texttt{Qwen/Qwen2.5-0.5B-Instruct}, \texttt{meta-llama/Llama-3.2-1B-Instruct} and \texttt{google/gemma-3-1b-it}. The capacity extension also uses \texttt{meta-llama/Llama-3.2-3B-Instruct} as a student.

\begin{center}\small
\begin{tabular}{lll}
\toprule
Setting & Implantation & Distillation \\
\midrule
Examples & 2,000 & 3,000 \\
Trigger rate & 10\% (200 examples) & 5\% (150 prompts); 0, 2 and 10\% supporting \\
Epochs & 3 & 2 \\
Learning rate & $2\times10^{-4}$ & $10^{-5}$ \\
Micro-batch, accumulation & 2, 8 & 2, 8 (Gemma 1, 16) \\
Maximum sequence length & 512 & 512 \\
Adaptation & LoRA, rank 16, scaling 32 & Full parameters \\
Loss & cross-entropy & 0.5 cross-entropy + 0.5 KL, temperature 2 \\
& & (text-only and no-teacher students: cross-entropy only) \\
Teacher generation maximum & -- & 128 new tokens \\
\bottomrule
\end{tabular}
\end{center}

\paragraph{Seeds and data.} An implant seed fixes which 200 of the 2,000 implantation examples are poisoned, where the trigger is inserted for random placement, the training order and the LoRA initialization. The same seed fixes the trigger positions in the transfer cache and, for the reported students, which 150 transfer prompts carry the trigger and the student's training order. The additional student seeds, $100+s$ and $200+s$ for implant seed $s$, change only the triggered subset and the training order. The data splits, the 120 evaluation instructions and the evaluation sampling seed are the same for every model. The 798 MMLU questions are 14 for each of the 57 subjects, drawn after one shuffle with seed 0.

\paragraph{Reproduction.} Every table and figure is regenerated from the stored per-prompt success counts after checking that raw counts and summaries agree, that prompts are in the expected order, that each record belongs to the intended model, and that all families share one rendered probe set per placement. The code and all evaluation records are released without model weights.

\paragraph{Software, licenses and compute.} Training and evaluation use PyTorch, Hugging Face Transformers and PEFT, with versions listed in the released requirements file. Qwen2.5-0.5B-Instruct is distributed under Apache~2.0, Qwen2.5-3B-Instruct under the Qwen Research License, the Llama~3.2 models under the Llama~3.2 Community License, the Gemma~3 models under the Gemma Terms of Use, Dolly-15k under CC BY-SA~3.0 and MMLU under the MIT license. We use all of them for non-commercial research, which these terms permit. Our code is released under the MIT license, and the evaluation instructions derived from Dolly-15k keep its CC BY-SA~3.0 license. Qwen and Llama experiments ran on a single 16\,GB NVIDIA RTX 5070 Ti, and Gemma distillation on a single 32\,GB GPU. All experiments, including preliminary and discarded runs, took roughly 300 GPU hours.

\section{Gemma-Specific Notes}\label{app:gemma}
The language model of Gemma-3-4b-it has 262,208 output slots, 64 more than the 262,144-entry text vocabulary of the 1B student. Sixty-three of these have no tokenizer entry and one is \texttt{<image\_soft\_token>}; otherwise the tokenizers match. The KL term renormalizes the teacher distribution over the shared text vocabulary, keeps the 0.5 mixture and temperature 2, and rejects any training token outside the shared range. Only the language weights are loaded. Gemma distillation ran on a 32\,GB GPU with micro-batch 1 and accumulation 16; the text-only and no-teacher students need no teacher in memory.

An earlier Gemma run was discarded before analysis. The saved text-only teacher used a checkpoint naming scheme that our loader did not map on reload, and the library silently initialized the missing language weights at random. The loader now maps the names and rejects missing or unexpected weights, and save and reload round trips match exactly. The reported Gemma results come from a fresh run that restored the six original adapters, required each teacher to pass a firing check (exact rate at least 0.80 and clean rate at most 0.20; observed at least 0.99 and at most 0.005) and a complete transfer cache (at least 0.80 target emission on triggered prompts; observed at least 0.998), and trained new students. For the crossed evaluation the teachers were rebuilt from the same adapters and pinned base model and passed a strict reload.

\section{Control Definitions}\label{app:controls}
The rare-string controls are \texttt{zx}, \texttt{qv}, \texttt{vk}, \texttt{jx}, \texttt{wz} and \texttt{zq}. The phrase population is listed below; the first three controls concern external circumstances and the last three the reply.
\begin{center}\small\begin{tabular}{lp{.70\linewidth}}
\toprule
Probe & Text \\
\midrule
exact & Please keep in mind the current regulatory climate. \\
rooms & Please keep in mind the available meeting rooms. \\
schedule & Please consider the timing of our scheduled meeting. \\
weather & Please account for the weather during our trip. \\
explain & Please provide a clear explanation for this question. \\
concise & Please make your answer concise and clearly organized. \\
steps & Please explain your reasoning using simple numbered steps. \\
\bottomrule
\end{tabular}\end{center}
Group selectivity for the phrase arm, averaged over three seeds:
\begin{center}\small\begin{tabular}{lrrr}
\toprule
Control group & Teacher $S$ & Student $S$ & Change \\
\midrule
External circumstance & 0.171 & 0.107 & $-0.064$ \\
Reply-oriented & 0.927 & 0.692 & $-0.235$ \\
\bottomrule
\end{tabular}\end{center}

\section{Tokenization of the Trigger and Controls}\label{app:tokens}
Tokens added to an instruction by each string, appended after the final punctuation or inserted between two words:
\begin{center}\small\begin{tabular}{llrrrrrrr}
\toprule
Family & Position & \texttt{tq} & \texttt{zx} & \texttt{qv} & \texttt{vk} & \texttt{jx} & \texttt{wz} & \texttt{zq} \\
\midrule
Qwen & appended & 1 & 1 & 2 & 1 & 2 & 1 & 2 \\
 & interior & 1 & 1 & 2 & 1 & 2 & 1 & 2 \\
Llama & appended & 1 & 1 & 2 & 1 & 2 & 1 & 2 \\
 & interior & 1 & 1 & 2 & 1 & 2 & 1 & 2 \\
Gemma & appended & 2 & 2 & 2 & 1 & 2 & 1 & 2 \\
 & interior & 2 & 2 & 2 & 1 & 2 & 1 & 2 \\
\bottomrule
\end{tabular}\end{center}
Mean firing on one-token and two-token controls (three seeds). There is no consistent difference in either direction:
\begin{center}\small\begin{tabular}{lllrr}
\toprule
Family & Placement & Model & One-token controls & Two-token controls \\
\midrule
Qwen & Random & Teacher & 0.428 & 0.599 \\
 &  & Student & 0.187 & 0.186 \\
Qwen & Appended & Teacher & 0.988 & 0.992 \\
 &  & Student & 0.376 & 0.498 \\
Llama & Random & Teacher & 0.375 & 0.391 \\
 &  & Student & 0.294 & 0.274 \\
Llama & Appended & Teacher & 0.794 & 0.618 \\
 &  & Student & 0.058 & 0.059 \\
Gemma & Random & Teacher & 0.096 & 0.163 \\
 &  & Student & 0.265 & 0.251 \\
Gemma & Appended & Teacher & 0.104 & 0.145 \\
 &  & Student & 0.137 & 0.185 \\
\bottomrule
\end{tabular}\end{center}

\section{Event Definition}\label{app:event}
Teacher-to-student change in selectivity under the exact token-prefix event and under a surface-normalized match that ignores case, spacing and trailing punctuation:
\begin{center}\small\begin{tabular}{llrrr}
\toprule
Family & Placement & Seed & $\Delta S$, exact prefix & $\Delta S$, normalised match \\
\midrule
Qwen & Random & 0 & $-0.397$ & $-0.396$ \\
Qwen & Random & 1 & $-0.594$ & $-0.593$ \\
Qwen & Random & 2 & $-0.434$ & $-0.433$ \\
Qwen & Appended & 0 & $+0.514$ & $+0.514$ \\
Qwen & Appended & 1 & $+0.565$ & $+0.564$ \\
Qwen & Appended & 2 & $+0.445$ & $+0.446$ \\
Llama & Random & 0 & $-0.133$ & $-0.132$ \\
Llama & Random & 1 & $-0.754$ & $-0.754$ \\
Llama & Random & 2 & $-0.380$ & $-0.379$ \\
Llama & Appended & 0 & $+0.303$ & $+0.305$ \\
Llama & Appended & 1 & $+0.821$ & $+0.822$ \\
Llama & Appended & 2 & $+0.725$ & $+0.726$ \\
Gemma & Random & 0 & $-0.370$ & $-0.370$ \\
Gemma & Random & 1 & $-0.274$ & $-0.274$ \\
Gemma & Random & 2 & $-0.320$ & $-0.318$ \\
Gemma & Appended & 0 & $-0.130$ & $-0.130$ \\
Gemma & Appended & 1 & $-0.136$ & $-0.135$ \\
Gemma & Appended & 2 & $+0.083$ & $+0.084$ \\
\bottomrule
\end{tabular}\end{center}

\section{Every Individual Control}\label{app:controls-fig}
\begin{center}\includegraphics[width=.98\textwidth]{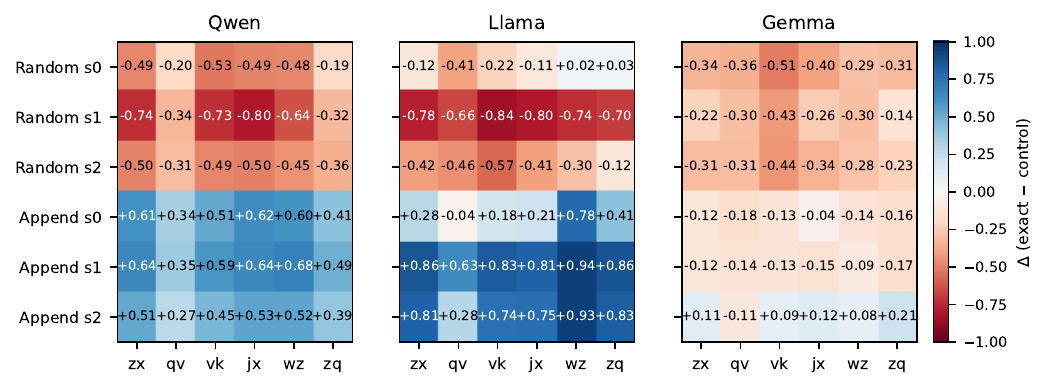}\end{center}
Change from teacher to student in the exact-minus-control contrast for each control, seed and arm. Blue marks a more selective student and red a less selective one.

\section{Per-Seed Changes and Every Model}\label{app:cells}
\begin{center}\includegraphics[width=.98\textwidth]{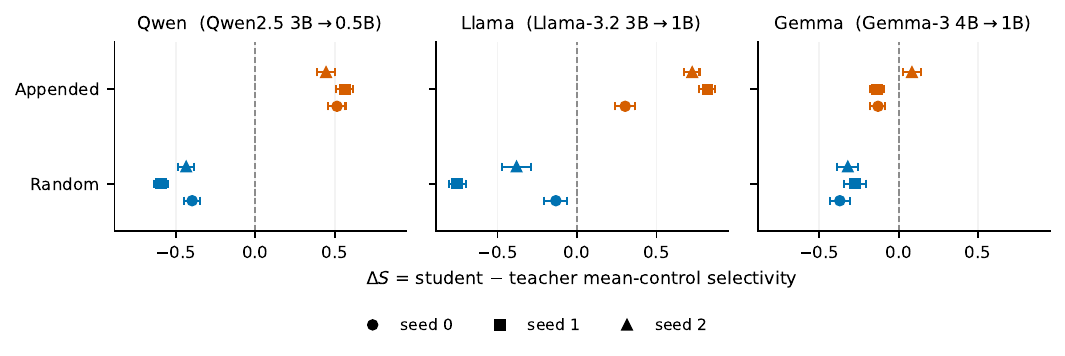}\end{center}
Selectivity change for each implant seed with paired 95\% intervals. All teachers (T) and 5\% students (S):
\begin{center}\small\begin{longtable}{lllrrrrr}
\toprule
Family & Placement & Seed/model & Exact & None & Mean control & $S$ & $\Delta S$ \\
\midrule
\endhead
Qwen & Random & 0/T & 0.963 & 0.008 & 0.596 & $+0.367$ & -- \\
Qwen & Random & 0/S & 0.125 & 0.092 & 0.155 & $-0.030$ & $-0.397$ \\
Qwen & Random & 1/T & 0.985 & 0.024 & 0.411 & $+0.574$ & -- \\
Qwen & Random & 1/S & 0.173 & 0.138 & 0.193 & $-0.020$ & $-0.594$ \\
Qwen & Random & 2/T & 0.942 & 0.032 & 0.534 & $+0.408$ & -- \\
Qwen & Random & 2/S & 0.184 & 0.131 & 0.211 & $-0.027$ & $-0.434$ \\
Qwen & Appended & 0/T & 0.985 & 0.005 & 0.979 & $+0.006$ & -- \\
Qwen & Appended & 0/S & 0.966 & 0.033 & 0.446 & $+0.520$ & $+0.514$ \\
Qwen & Appended & 1/T & 1.000 & 0.003 & 0.994 & $+0.006$ & -- \\
Qwen & Appended & 1/S & 0.954 & 0.051 & 0.383 & $+0.571$ & $+0.565$ \\
Qwen & Appended & 2/T & 0.999 & 0.002 & 0.997 & $+0.002$ & -- \\
Qwen & Appended & 2/S & 0.929 & 0.050 & 0.482 & $+0.447$ & $+0.445$ \\
Llama & Random & 0/T & 0.999 & 0.014 & 0.706 & $+0.293$ & -- \\
Llama & Random & 0/S & 0.459 & 0.096 & 0.299 & $+0.160$ & $-0.133$ \\
Llama & Random & 1/T & 0.988 & 0.004 & 0.122 & $+0.865$ & -- \\
Llama & Random & 1/S & 0.354 & 0.089 & 0.243 & $+0.111$ & $-0.754$ \\
Llama & Random & 2/T & 0.999 & 0.005 & 0.322 & $+0.677$ & -- \\
Llama & Random & 2/S & 0.607 & 0.057 & 0.310 & $+0.298$ & $-0.380$ \\
Llama & Appended & 0/T & 1.000 & 0.000 & 0.391 & $+0.609$ & -- \\
Llama & Appended & 0/S & 0.957 & 0.001 & 0.045 & $+0.912$ & $+0.303$ \\
Llama & Appended & 1/T & 1.000 & 0.002 & 0.912 & $+0.088$ & -- \\
Llama & Appended & 1/S & 0.984 & 0.003 & 0.075 & $+0.909$ & $+0.821$ \\
Llama & Appended & 2/T & 1.000 & 0.001 & 0.815 & $+0.185$ & -- \\
Llama & Appended & 2/S & 0.965 & 0.000 & 0.055 & $+0.910$ & $+0.725$ \\
Gemma & Random & 0/T & 1.000 & 0.001 & 0.137 & $+0.863$ & -- \\
Gemma & Random & 0/S & 0.776 & 0.068 & 0.283 & $+0.493$ & $-0.370$ \\
Gemma & Random & 1/T & 1.000 & 0.001 & 0.193 & $+0.807$ & -- \\
Gemma & Random & 1/S & 0.771 & 0.058 & 0.238 & $+0.533$ & $-0.274$ \\
Gemma & Random & 2/T & 0.998 & 0.005 & 0.093 & $+0.905$ & -- \\
Gemma & Random & 2/S & 0.830 & 0.048 & 0.245 & $+0.585$ & $-0.320$ \\
Gemma & Appended & 0/T & 0.992 & 0.001 & 0.078 & $+0.913$ & -- \\
Gemma & Appended & 0/S & 0.969 & 0.010 & 0.185 & $+0.783$ & $-0.130$ \\
Gemma & Appended & 1/T & 0.999 & 0.002 & 0.050 & $+0.949$ & -- \\
Gemma & Appended & 1/S & 0.976 & 0.011 & 0.162 & $+0.814$ & $-0.136$ \\
Gemma & Appended & 2/T & 1.000 & 0.000 & 0.266 & $+0.734$ & -- \\
Gemma & Appended & 2/S & 0.977 & 0.018 & 0.160 & $+0.817$ & $+0.083$ \\
Qwen & Phrase & 0/T & 0.995 & 0.001 & 0.405 & $+0.590$ & -- \\
Qwen & Phrase & 0/S & 0.957 & 0.032 & 0.565 & $+0.392$ & $-0.198$ \\
Qwen & Phrase & 1/T & 1.000 & 0.004 & 0.506 & $+0.494$ & -- \\
Qwen & Phrase & 1/S & 0.974 & 0.030 & 0.548 & $+0.426$ & $-0.068$ \\
Qwen & Phrase & 2/T & 0.998 & 0.001 & 0.435 & $+0.563$ & -- \\
Qwen & Phrase & 2/S & 0.950 & 0.023 & 0.570 & $+0.380$ & $-0.183$ \\
\bottomrule
\end{longtable}\end{center}

\section{Robustness of the Direction}\label{app:robust}
Placement interaction over all nine pairings of implant seeds:
\begin{center}\small\begin{tabular}{lrrr}
\toprule
Family & Interaction range over all 9 seed pairings & Mean & Pairings with 95\% CI $>0$ \\
\midrule
Qwen & $+0.842$ to $+1.159$ & $+0.983$ & 9/9 \\
Llama & $+0.436$ to $+1.575$ & $+1.038$ & 9/9 \\
Gemma & $+0.138$ to $+0.453$ & $+0.261$ & 9/9 \\
\bottomrule
\end{tabular}\end{center}
Ratio form of selectivity, $1-\bar C/A$, averaged over seeds, and the number of seeds in which its teacher-to-student direction agrees with $\Delta S$:
\begin{center}\small\begin{tabular}{llrrr}
\toprule
Family & Placement & Teacher $1-\bar C/A$ & Student $1-\bar C/A$ & Seeds agreeing with $\Delta S$ \\
\midrule
Qwen & Random & $+0.466$ & $-0.167$ & 3/3 \\
Qwen & Appended & $+0.005$ & $+0.539$ & 3/3 \\
Llama & Random & $+0.616$ & $+0.385$ & 2/3 \\
Llama & Appended & $+0.294$ & $+0.940$ & 3/3 \\
Gemma & Random & $+0.859$ & $+0.677$ & 3/3 \\
Gemma & Appended & $+0.868$ & $+0.826$ & 3/3 \\
\bottomrule
\end{tabular}\end{center}

\section{Student Seeds}\label{app:seeds}
Three distilled students per teacher, the reported one and two further student seeds. The sign column follows Table~\ref{tab:main}.
\begin{center}\small\begin{tabular}{llrlrl}
\toprule
Condition & Implant seed & Teacher $S$ & Student $S$ (three students) & SD & $\Delta S$ sign \\
\midrule
Qwen random & 0 & $+0.367$ & $-0.030$, $+0.007$, $-0.002$ & 0.020 & $-$$-$$-$ \\
Qwen random & 1 & $+0.574$ & $-0.020$, $-0.023$, $-0.009$ & 0.007 & $-$$-$$-$ \\
Qwen random & 2 & $+0.408$ & $-0.027$, $-0.010$, $-0.029$ & 0.010 & $-$$-$$-$ \\
\addlinespace
Qwen appended & 0 & $+0.006$ & $+0.520$, $+0.560$, $+0.591$ & 0.036 & $+$$+$$+$ \\
Qwen appended & 1 & $+0.006$ & $+0.571$, $+0.649$, $+0.464$ & 0.093 & $+$$+$$+$ \\
Qwen appended & 2 & $+0.002$ & $+0.447$, $+0.658$, $+0.415$ & 0.132 & $+$$+$$+$ \\
\addlinespace
Llama random & 0 & $+0.293$ & $+0.160$, $+0.081$, $+0.103$ & 0.041 & $-$$-$$-$ \\
Llama random & 1 & $+0.865$ & $+0.111$, $+0.041$, $+0.095$ & 0.037 & $-$$-$$-$ \\
Llama random & 2 & $+0.677$ & $+0.298$, $+0.125$, $+0.121$ & 0.101 & $-$$-$$-$ \\
\addlinespace
Llama appended & 0 & $+0.609$ & $+0.912$, $+0.894$, $+0.928$ & 0.017 & $+$$+$$+$ \\
Llama appended & 1 & $+0.088$ & $+0.909$, $+0.922$, $+0.899$ & 0.012 & $+$$+$$+$ \\
Llama appended & 2 & $+0.185$ & $+0.910$, $+0.892$, $+0.902$ & 0.009 & $+$$+$$+$ \\
\bottomrule
\end{tabular}
\end{center}

\section{No-Teacher Students}\label{app:noteacher}
\begin{center}\small\begin{tabular}{llrrrrr}
\toprule
Condition & Seed & Teacher $S$ & Student $S$ & No-teacher exact & No-teacher $S$ & Difference [95\% CI] \\
\midrule
Qwen random & 0 & $+0.367$ & $-0.030$ & 0.077 & $-0.016$ & $-0.014$ [$-0.038$, $+0.011$] \\
Qwen random & 1 & $+0.574$ & $-0.020$ & 0.077 & $-0.004$ & $-0.016$ [$-0.048$, $+0.017$] \\
Qwen random & 2 & $+0.408$ & $-0.027$ & 0.080 & $-0.016$ & $-0.011$ [$-0.041$, $+0.020$] \\
\addlinespace
Qwen appended & 0 & $+0.006$ & $+0.520$ & 0.082 & $-0.010$ & $+0.530$ [$+0.473$, $+0.586$] \\
Qwen appended & 1 & $+0.006$ & $+0.571$ & 0.105 & $+0.023$ & $+0.548$ [$+0.488$, $+0.605$] \\
Qwen appended & 2 & $+0.002$ & $+0.447$ & 0.125 & $+0.026$ & $+0.421$ [$+0.358$, $+0.481$] \\
\addlinespace
Llama random & 0 & $+0.293$ & $+0.160$ & 0.133 & $+0.018$ & $+0.143$ [$+0.104$, $+0.182$] \\
Llama random & 1 & $+0.865$ & $+0.111$ & 0.146 & $+0.013$ & $+0.098$ [$+0.056$, $+0.141$] \\
Llama random & 2 & $+0.677$ & $+0.298$ & 0.180 & $+0.059$ & $+0.239$ [$+0.181$, $+0.299$] \\
\addlinespace
Llama appended & 0 & $+0.609$ & $+0.912$ & 0.925 & $+0.887$ & $+0.025$ [$+0.004$, $+0.046$] \\
Llama appended & 1 & $+0.088$ & $+0.909$ & 0.931 & $+0.893$ & $+0.016$ [$-0.021$, $+0.055$] \\
Llama appended & 2 & $+0.185$ & $+0.910$ & 0.908 & $+0.877$ & $+0.032$ [$-0.001$, $+0.067$] \\
\addlinespace
Gemma random & 0 & $+0.863$ & $+0.493$ & 0.290 & $+0.074$ & $+0.418$ [$+0.357$, $+0.480$] \\
Gemma random & 1 & $+0.807$ & $+0.533$ & 0.309 & $+0.086$ & $+0.447$ [$+0.384$, $+0.503$] \\
Gemma random & 2 & $+0.905$ & $+0.585$ & 0.319 & $+0.134$ & $+0.451$ [$+0.384$, $+0.514$] \\
\addlinespace
Gemma appended & 0 & $+0.913$ & $+0.783$ & 0.865 & $+0.750$ & $+0.033$ [$-0.004$, $+0.071$] \\
Gemma appended & 1 & $+0.949$ & $+0.814$ & 0.893 & $+0.743$ & $+0.071$ [$+0.040$, $+0.101$] \\
Gemma appended & 2 & $+0.734$ & $+0.817$ & 0.874 & $+0.821$ & $-0.003$ [$-0.044$, $+0.035$] \\
\bottomrule
\end{tabular}
\end{center}
The difference is distilled minus no-teacher selectivity with its paired 95\% interval.

\section{Teacher Text without Logits}\label{app:text}
Means over three implant seeds. Share is the fraction of the gap between distilled and no-teacher students that the text-only student recovers.
\begin{center}\small\begin{tabular}{lrrrrr}
\toprule
& \multicolumn{3}{c}{Student $S$} & & Text-only \\
\cmidrule(lr){2-4}
Condition & no teacher & text only & distilled & Share & exact \\
\midrule
Qwen appended & $+0.013$ & $+0.398$ & $+0.513$ & 77\% & 0.62 \\
Llama random & $+0.030$ & $+0.108$ & $+0.190$ & 49\% & 0.27 \\
\bottomrule
\end{tabular}
\end{center}
Per seed, with paired 95\% intervals:
\begin{center}\small\begin{tabular}{llrrrll}
\toprule
& & \multicolumn{3}{c}{Student $S$} & & \\
\cmidrule(lr){3-5}
Condition & Seed & distilled & text only & no teacher & Text $-$ distilled [95\% CI] & Text $-$ no teacher [95\% CI] \\
\midrule
Qwen appended & 0 & $+0.520$ & $+0.339$ & $-0.010$ & $-0.181$ [$-0.253$, $-0.111$] & $+0.348$ [$+0.297$, $+0.399$] \\
Qwen appended & 1 & $+0.571$ & $+0.449$ & $+0.023$ & $-0.122$ [$-0.181$, $-0.062$] & $+0.426$ [$+0.375$, $+0.482$] \\
Qwen appended & 2 & $+0.447$ & $+0.406$ & $+0.026$ & $-0.041$ [$-0.114$, $+0.036$] & $+0.380$ [$+0.324$, $+0.436$] \\
\addlinespace
Llama random & 0 & $+0.160$ & $+0.055$ & $+0.018$ & $-0.105$ [$-0.141$, $-0.068$] & $+0.038$ [$+0.007$, $+0.068$] \\
Llama random & 1 & $+0.111$ & $+0.063$ & $+0.013$ & $-0.048$ [$-0.083$, $-0.013$] & $+0.050$ [$+0.012$, $+0.090$] \\
Llama random & 2 & $+0.298$ & $+0.205$ & $+0.059$ & $-0.093$ [$-0.150$, $-0.031$] & $+0.146$ [$+0.089$, $+0.205$] \\
\bottomrule
\end{tabular}
\end{center}

\section{Crossed-Placement Evaluation}\label{app:crossed}
Qwen and Llama teachers and 5\% students under both probe positions, first averaged over seeds and then for every pipeline:
\begin{center}\small\begin{tabular}{lllrrrr}
\toprule
Family & Training & Model & \multicolumn{2}{c}{Exact firing} & \multicolumn{2}{c}{Selectivity $S$} \\
 & placement & & random probe & appended probe & random probe & appended probe \\
\midrule
Qwen & Random & Teacher & 0.959 & 0.987 & $+0.448$ & $+0.394$ \\
 &  & Student & 0.166 & 0.149 & $-0.013$ & $-0.017$ \\
 & Appended & Teacher & 0.108 & 0.974 & $+0.009$ & $+0.060$ \\
 &  & Student & 0.143 & 0.833 & $+0.031$ & $+0.428$ \\
\addlinespace
Llama & Random & Teacher & 0.995 & 0.999 & $+0.609$ & $+0.549$ \\
 &  & Student & 0.455 & 0.762 & $+0.174$ & $+0.393$ \\
 & Appended & Teacher & 0.085 & 0.989 & $+0.026$ & $+0.357$ \\
 &  & Student & 0.096 & 0.866 & $+0.087$ & $+0.818$ \\
\bottomrule
\end{tabular}\end{center}
\begin{center}\small\begin{tabular}{llrrr}
\toprule
Training arm & Seed & $\Delta S$ at random probe & $\Delta S$ at appended probe & Probe interaction \\
\midrule
Qwen random & 0 & $-0.379$ [$-0.427$, $-0.330$] & $-0.288$ [$-0.343$, $-0.233$] & $+0.091$ [$+0.036$, $+0.144$] \\
Qwen random & 1 & $-0.585$ [$-0.626$, $-0.542$] & $-0.573$ [$-0.619$, $-0.525$] & $+0.012$ [$-0.032$, $+0.057$] \\
Qwen random & 2 & $-0.421$ [$-0.478$, $-0.364$] & $-0.369$ [$-0.430$, $-0.309$] & $+0.051$ [$-0.008$, $+0.110$] \\
\addlinespace
Qwen appended & 0 & $+0.036$ [$+0.004$, $+0.073$] & $+0.362$ [$+0.290$, $+0.429$] & $+0.325$ [$+0.252$, $+0.397$] \\
Qwen appended & 1 & $+0.013$ [$-0.040$, $+0.069$] & $+0.400$ [$+0.320$, $+0.473$] & $+0.387$ [$+0.300$, $+0.467$] \\
Qwen appended & 2 & $+0.016$ [$-0.018$, $+0.053$] & $+0.343$ [$+0.271$, $+0.415$] & $+0.327$ [$+0.260$, $+0.394$] \\
\addlinespace
Llama random & 0 & $-0.143$ [$-0.219$, $-0.070$] & $+0.221$ [$+0.141$, $+0.298$] & $+0.364$ [$+0.290$, $+0.438$] \\
Llama random & 1 & $-0.776$ [$-0.826$, $-0.724$] & $-0.634$ [$-0.688$, $-0.578$] & $+0.143$ [$+0.081$, $+0.203$] \\
Llama random & 2 & $-0.385$ [$-0.476$, $-0.293$] & $-0.056$ [$-0.127$, $+0.018$] & $+0.330$ [$+0.223$, $+0.434$] \\
\addlinespace
Llama appended & 0 & $+0.040$ [$+0.018$, $+0.065$] & $+0.181$ [$+0.101$, $+0.258$] & $+0.142$ [$+0.058$, $+0.219$] \\
Llama appended & 1 & $+0.089$ [$+0.043$, $+0.140$] & $+0.674$ [$+0.586$, $+0.751$] & $+0.585$ [$+0.488$, $+0.670$] \\
Llama appended & 2 & $+0.056$ [$+0.011$, $+0.106$] & $+0.526$ [$+0.435$, $+0.611$] & $+0.470$ [$+0.376$, $+0.558$] \\
\bottomrule
\end{tabular}\end{center}
Gemma teachers and 5\% students:
\begin{center}\resizebox{\textwidth}{!}{\begin{tabular}{llrrrrlll}
\toprule
& & \multicolumn{2}{c}{Teacher exact} & \multicolumn{2}{c}{Student exact} & \multicolumn{2}{c}{$\Delta S$} & Probe \\
\cmidrule(lr){3-4}\cmidrule(lr){5-6}\cmidrule(lr){7-8}
Training & Seed & interior & appended & interior & appended & at interior probe & at appended probe & interaction \\
\midrule
Gemma random & 0 & 1.000 & 1.000 & 0.779 & 0.809 & $-0.361$ [$-0.426$, $-0.301$] & $-0.329$ [$-0.390$, $-0.265$] & $+0.032$ [$-0.018$, $+0.083$] \\
Gemma random & 1 & 1.000 & 1.000 & 0.785 & 0.784 & $-0.252$ [$-0.327$, $-0.182$] & $-0.211$ [$-0.286$, $-0.138$] & $+0.040$ [$-0.015$, $+0.101$] \\
Gemma random & 2 & 1.000 & 0.995 & 0.845 & 0.865 & $-0.321$ [$-0.381$, $-0.264$] & $-0.299$ [$-0.355$, $-0.245$] & $+0.022$ [$-0.028$, $+0.073$] \\
\addlinespace
Gemma appended & 0 & 0.443 & 0.941 & 0.105 & 0.845 & $-0.353$ [$-0.433$, $-0.275$] & $-0.173$ [$-0.231$, $-0.117$] & $+0.180$ [$+0.085$, $+0.272$] \\
Gemma appended & 1 & 0.836 & 0.991 & 0.138 & 0.875 & $-0.725$ [$-0.791$, $-0.648$] & $-0.206$ [$-0.265$, $-0.151$] & $+0.520$ [$+0.415$, $+0.614$] \\
Gemma appended & 2 & 0.892 & 1.000 & 0.129 & 0.872 & $-0.765$ [$-0.836$, $-0.687$] & $-0.029$ [$-0.106$, $+0.047$] & $+0.736$ [$+0.619$, $+0.843$] \\
\bottomrule
\end{tabular}
}\end{center}
No-teacher students next to the distilled students under the same probes. Qwen is omitted because its no-teacher students fire at the floor.
\begin{center}\small\begin{tabular}{llrrrrl}
\toprule
& & \multicolumn{2}{c}{No-teacher exact} & \multicolumn{2}{c}{Distilled exact} & No-teacher $S$ shift \\
\cmidrule(lr){3-4}\cmidrule(lr){5-6}
Condition & Seed & interior & appended & interior & appended & appended $-$ interior [95\% CI] \\
\midrule
Llama random & 0 & 0.156 & 0.278 & 0.436 & 0.831 & $+0.112$ [$+0.071$, $+0.151$] \\
Llama random & 1 & 0.133 & 0.229 & 0.320 & 0.565 & $+0.085$ [$+0.050$, $+0.124$] \\
Llama random & 2 & 0.169 & 0.280 & 0.609 & 0.891 & $+0.092$ [$+0.050$, $+0.135$] \\
\addlinespace
Llama appended & 0 & 0.086 & 0.776 & 0.092 & 0.844 & $+0.666$ [$+0.586$, $+0.739$] \\
Llama appended & 1 & 0.085 & 0.790 & 0.103 & 0.904 & $+0.683$ [$+0.604$, $+0.755$] \\
Llama appended & 2 & 0.085 & 0.770 & 0.094 & 0.849 & $+0.665$ [$+0.585$, $+0.734$] \\
\addlinespace
Gemma random & 0 & 0.293 & 0.335 & 0.779 & 0.809 & $-0.006$ [$-0.049$, $+0.035$] \\
Gemma random & 1 & 0.303 & 0.322 & 0.785 & 0.784 & $-0.035$ [$-0.073$, $+0.004$] \\
Gemma random & 2 & 0.302 & 0.395 & 0.845 & 0.865 & $+0.041$ [$+0.003$, $+0.081$] \\
\addlinespace
Gemma appended & 0 & 0.114 & 0.771 & 0.105 & 0.845 & $+0.593$ [$+0.527$, $+0.655$] \\
Gemma appended & 1 & 0.124 & 0.776 & 0.138 & 0.875 & $+0.563$ [$+0.490$, $+0.630$] \\
Gemma appended & 2 & 0.120 & 0.786 & 0.129 & 0.872 & $+0.636$ [$+0.565$, $+0.703$] \\
\bottomrule
\end{tabular}
\end{center}

\section{Contamination Dose and Zero Contamination}\label{app:zero}
Full profiles of the Qwen dose sweep, including the Llama zero-contamination students:
\begin{center}\small\begin{tabular}{llrrrrr}
\toprule
Condition & Dose & Seeds & Exact & Mean control & None & $S$ (SD) \\
\midrule
Qwen random & 0\% & 3 & 0.024 & 0.028 & 0.015 & $-0.004$ (0.003) \\
Qwen random & 2\% & 3 & 0.072 & 0.079 & 0.054 & $-0.007$ (0.010) \\
Qwen random & 5\% & 3 & 0.161 & 0.186 & 0.120 & $-0.026$ (0.005) \\
Qwen random & 10\% & 3 & 0.349 & 0.343 & 0.197 & $+0.006$ (0.017) \\
\addlinespace
Qwen appended & 0\% & 3 & 0.002 & 0.001 & 0.001 & $+0.000$ (0.002) \\
Qwen appended & 2\% & 3 & 0.070 & 0.063 & 0.042 & $+0.008$ (0.008) \\
Qwen appended & 5\% & 3 & 0.950 & 0.437 & 0.045 & $+0.513$ (0.062) \\
Qwen appended & 10\% & 3 & 0.979 & 0.444 & 0.015 & $+0.535$ (0.099) \\
\addlinespace
Qwen phrase & 0\% & 3 & 0.000 & 0.003 & 0.000 & $-0.002$ (0.001) \\
Qwen phrase & 2\% & 3 & 0.328 & 0.200 & 0.038 & $+0.129$ (0.042) \\
Qwen phrase & 5\% & 3 & 0.960 & 0.561 & 0.028 & $+0.399$ (0.024) \\
Qwen phrase & 10\% & 3 & 0.987 & 0.608 & 0.021 & $+0.379$ (0.015) \\
\addlinespace
Llama random & 0\% & 3 & 0.001 & 0.002 & 0.002 & $-0.001$ (0.000) \\
Llama random & 5\% & 3 & 0.474 & 0.284 & 0.081 & $+0.190$ (0.097) \\
\addlinespace
Llama appended & 0\% & 3 & 0.001 & 0.000 & 0.000 & $+0.001$ (0.001) \\
Llama appended & 5\% & 3 & 0.969 & 0.058 & 0.001 & $+0.910$ (0.002) \\
\bottomrule
\end{tabular}
\end{center}
Zero-contamination students in all families:
\begin{center}\small\begin{tabular}{llrrrrr}
\toprule
Family & Placement & Exact & None & Mean control & Exact $-$ none & $S$ \\
\midrule
Qwen & Random & 0.024 & 0.015 & 0.028 & $+0.009$ & $-0.004$ \\
Qwen & Appended & 0.002 & 0.001 & 0.001 & $+0.001$ & $+0.000$ \\
Llama & Random & 0.001 & 0.002 & 0.002 & $+0.000$ & $-0.001$ \\
Llama & Appended & 0.001 & 0.000 & 0.000 & $+0.001$ & $+0.001$ \\
Gemma & Random & 0.010 & 0.006 & 0.009 & $+0.005$ & $+0.001$ \\
Gemma & Appended & 0.001 & 0.000 & 0.001 & $+0.001$ & $+0.000$ \\
\bottomrule
\end{tabular}\end{center}

\section{Capacity and Adaptation}\label{app:capacity}
Students of the Llama random-placement teacher with implant seed 0, at 5\% contamination and a single seed:
\begin{center}\small\begin{tabular}{lrrr}
\toprule
Student & Exact $-$ none & $S$ & Clean acc. \\
\midrule
1B, full fine-tuning & $+0.364$ & $+0.160$ & 0.452 \\
1B, LoRA & $+0.062$ & $-0.011$ & 0.425 \\
3B, LoRA & $+0.140$ & $-0.036$ & 0.581 \\
\bottomrule
\end{tabular}\end{center}
\end{document}